\documentclass{article}
\usepackage{spconf,amsmath,amssymb,amsfonts,graphicx}
\usepackage{subfigure,lineno,multirow,booktabs,diagbox}
\usepackage[colorlinks,linkcolor=red]{hyperref}

\title{P$^2$-GAN: Efficient Style Transfer using Single Style Image}
\name{Zhentan Zheng$^{\dag}$ , Jianyi Liu$^{\dag}$ \sthanks{Corresponding author: jyliu@xjtu.edu.cn. This work was supported by the National Natural Science Foundation
of China (No.61379104) and the Provincial Key Laboratory
Program of Shaanxi (No. 2013SZS12-K01).}}
\address{$^{\dag}$ Institute
 of Artificial Intelligence and Robotics, Xi’an Jiaotong University,
 Shaanxi 710049, China}
\begin{document}
%
\maketitle
\begin{abstract}
Style transfer is a useful image synthesis technique that can re-render given image into another artistic style while preserving its content information. Generative Adversarial Network (GAN) is a widely adopted framework toward this task for its better representation ability on local style patterns than the traditional Gram-matrix based methods. However, most previous methods rely on sufficient amount of pre-collected style images to train the model. In this paper, a novel Patch Permutation GAN (P$^2$-GAN) network that can efficiently learn the stroke style from a single style image is proposed. We use patch permutation to generate multiple training samples from the given style image. A patch discriminator that can simultaneously process patch-wise images and natural images seamlessly is designed. We also propose a local texture descriptor based criterion to quantitatively evaluate the style transfer quality. Experimental results showed that our method can produce finer quality re-renderings from single style image with improved computational efficiency compared with many state-of-the-arts methods.
\end{abstract}
\section{Introduction}
\label{sec:intro}

Style transfer aims at redrawing a content image into another artistic style while preserving its semantic content. It has a rich history in terms of texture transfer \cite{efros2001image}\cite{ashikhmin2003fast}. Style transfer technique enables amateurs to produce fantastic and versatile images in seconds, so that it has many practical applications in cartoon generation, oil painting re-rendering and art education etc \cite{bib.tst.review}.

Most early works on texture transfer were pixel-level approaches that relied on low-level visual features. Since Gatys \cite{gatys2015a} for the first time incorporated convolutional neural networks (CNNs) into the computations of style loss and content loss, CNNs based approaches have become the mainstream of style transfer and achieved convincing results. These methods can be divided into three categories: a) online learning methods \cite{gatys} \cite{li2016precomputed}, b) patch-swap methods \cite{CNNMRF}\cite{DIA}\cite{DFR} , c) offline learning methods.

The former two kinds of methods usually suffer from expensive online computational time (20s\verb|~|300s) since no style prior information has been taken into consideration. On the contrary, offline learning methods always use the feed-forward networks, mostly auto-encoder,  trained from a set of pre-collected style images to memorize the  style information, so that the target image can be obtained in real time through feed-forward computation\cite{Johnson2016Perceptual}\cite{stylebank}\cite{stIn} . These methods can be regarded as a trade of computational time from online stage to offline stage.

\begin{figure}[t]

    \begin{minipage}[t]{0.24\linewidth}
    \includegraphics[width=2.08cm]{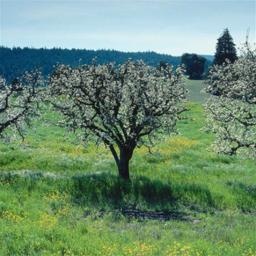}\vspace{2pt}
    \includegraphics[width=2.08cm]{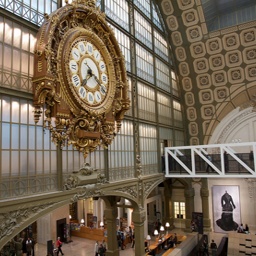}\vspace{2pt}
  \centerline{(a) Content \, }
  \centerline{ \ images}
    \end{minipage}
\hfill
    \begin{minipage}[t]{0.24\linewidth}
    \includegraphics[width=2.08cm]{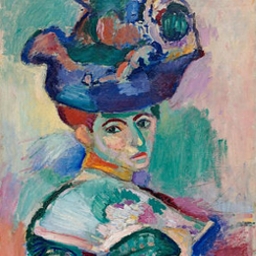}\vspace{2pt}
    \includegraphics[width=2.08cm]{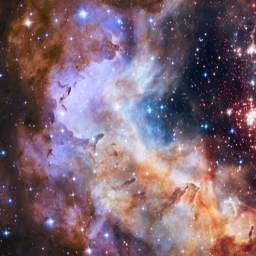}\vspace{2pt}
  \centerline{(b) Style \ }
  \centerline{ \, images}
    \end{minipage}
\hfill
    \begin{minipage}[t]{0.24\linewidth}
    \includegraphics[width=2.08cm]{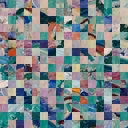}\vspace{2pt}
    \includegraphics[width=2.08cm]{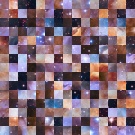}\vspace{2pt}
  \centerline{(c) Patch\quad}
  \centerline{Permutation}
    \end{minipage}
\hfill
    \begin{minipage}[t]{0.24\linewidth}
    \includegraphics[width=2.08cm]{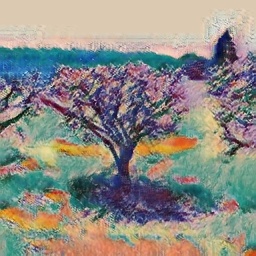}\vspace{2pt}
    \includegraphics[width=2.08cm]{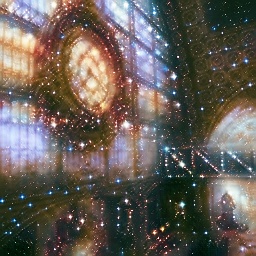}\vspace{2pt}
  {(d) Our results}
    \end{minipage}
    
    \vspace{-0.2cm}
  \caption{Style transfer from single style image by our method.}
    \vspace{-0.3cm}
  \label{fig-front}
\end{figure}

\begin{figure*}[t]
  \centering
  \includegraphics[width=16cm]{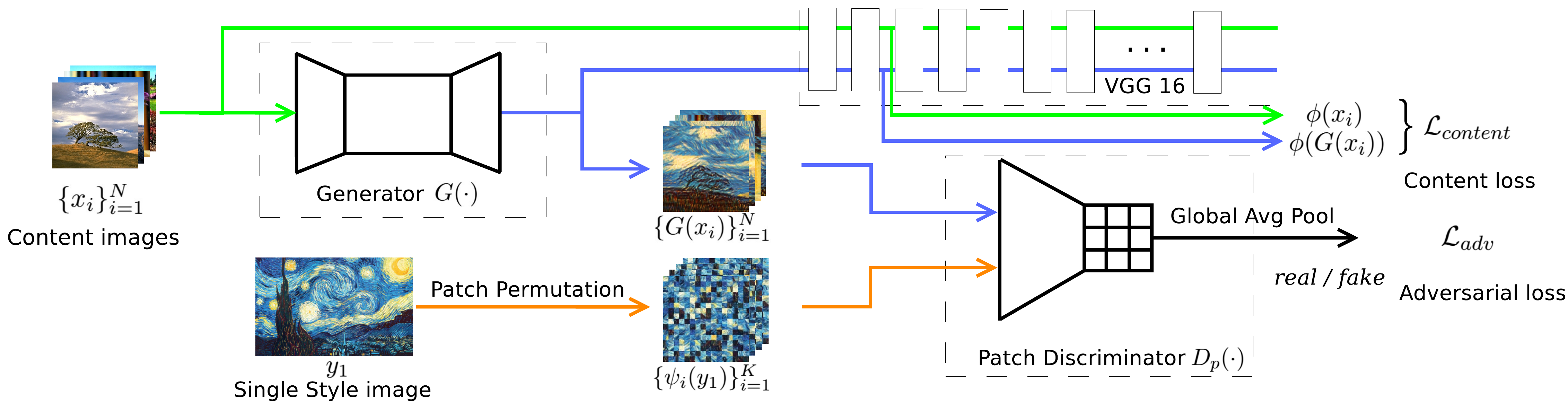}
  \caption{System overview and network architecture of the proposed P$^2$-GAN.}
  \label{fig::overview}
\end{figure*}

Generative Adversarial Networks (GANs) based style transfer methods \cite{cycle-gan}\cite{gan.gated} belong to the offline learning category. The generative network implicitly learns the target style, aiming to fool the discriminator. While the discriminative network try to distinguish the real/fake images by two-player minimax game. The \emph{adversarial loss} instead of the Gram-matrix based style loss can better learn the local style patterns. However, collecting sufficient amount of style images is sometimes difficult in practice and always makes the output style hard to control. That is, multiple training images are apt to generate unexpected texture effect since no two style images are exactly the same. 

In this paper, we propose to use only one style image for the training of GANs. It will make it possible to precisely simulate the expected stroke style from the customer-designated image without interfered by other sources. A novel Patch Permutation GAN (P$^2$-GAN) network is proposed. Our idea is to randomly break the unique style image into multiple patches which will be used as the training set. To the best of our knowledge, only one GAN-based previous work \cite{li2016precomputed} has taken into consideration of the single image training problem in style transfer. The main difference between MGAN \cite{li2016precomputed} and our method is that their patches locate in the feature maps while ours are segmented in the original image space for a better control of local stroke style. In addition, we have also designed a novel \emph{patch discriminator} to simultaneously process patch-wise images and natural images, which makes our network training more efficient. 

The rest of paper is organized as follows: In Section 2, our proposed P$^2$-GAN method will be described in detail. Then, we will discuss the experimental results in Section 3, and conclude in Section 4.

\section{Proposed Method}
\label{sec:method}

\subsection{Overview}
\label{sec::method::overview}
In most GAN-based style transfer methods \cite{ast}\cite{gan.gated}, a generator $G( \cdot )$ and a discriminator $D( \cdot )$ are usually learned from a set of content images $  \mathcal{X} =\{ x_i  \}_{i=1} ^N  $ and style images $ \mathcal{Y}  =\{ y_i  \}_{i=1} ^M  $ simultaneously, where $N$ and $M$ denote the numbers of content/style images respectively. Given a new content image $x_t$, the re-rendered image $G(x_t)$ will preserve the content information of $x_t$ and the style information of $\mathcal{Y}$.

When $ M=1 $, obviously traditional training methods of GANs will not work due to single sample can not represent a distribution, the network may even collapse while training. To overcome this problem, we have designed a novel network architecture as shown in Figure.\ref{fig::overview}. A patch permutation module and a patch discriminator $D_p(\cdot)$ are proposed, together with an improved encoder-decoder generator $G( \cdot )$ for more efficient computations on both offline and online stages.

\subsection{Patch Permutation}
\label{sec::method::permutation}

Since we only have one style image, we considered to break the style image into many patches, and learn stroke style from the patches. Therefore, a permutation method for the style image on the image was proposed. 

Firstly, patches with size $n \times n$ are cropped from style image $y_1$ at its random position, $n$ should be selected according to the scale of texture element in $y_1$. For any patch $p_j$, we have:

\vspace{-0.5cm}

\begin{equation}
  \label{eqn::randcrop}
  p_j(u, v) = y_1[(W_{y} - n) r + u, (H_{y} - n) r + v]
\end{equation}

\noindent
where $u$ and $v$ are the horizontal and vertical coordinates, $r$ is a random variable with uniform distribute ranging within [0, 1], $W_y$ and $H_y$ denote the width and height of image $y_1$ respectively. After $T^2$ times random cropping, all the patches $p_j, j = 1, ..., T^2$, can be reorganized into a new image $\psi(y_1)$ that yield to the following block-matrix form:

\vspace{-0.3cm}

\begin{equation}
  \label{eqn::pp}
	\psi(y_1) = \left[
	\begin{array}{cccc}
	p_{1}&p_{2}&\cdots&p_{T}\\
	p_{T+1}&p_{T+2}&\cdots&p_{2T}\\
	\vdots&\vdots&\ddots&\vdots\\
	p_{(T-1)T+1}&p_{(T-1)T + 2}&\cdots&p_{T^2}\\
	\end{array}
	\right]
\end{equation}

\noindent
Which $\psi(y_1) $ has a size of $nT \times nT$ and can be regarded as a \emph{permutation} of the original style image $y_1$.

Repeating such permutation $K$ times, the so obtained image set $\mathcal{Y}_{\psi} = \{ \psi_i(y_1)  \}_{i=1}^K$ will be used as the training set for our model instead of the single image $y_1$. In $\mathcal{Y}_{\psi}$, the stroke style information of $y_1$ will be preserved by selecting a suitable patch size $n$. Meanwhile, the structure information of $y_1$ has been discarded, which is benefit to avoid content interference from $y_1$.

\subsection{Patch Discriminator}
\label{sec::method::patch-d}

The operation of patch permutation in previous section will inevitably generate discontinuous textures on junction pixels between adjacent patches in $\psi_i(y_1)$. This will incur severe artifacts in the transferred image by traditional discriminators. To address this problem, we proposed a discriminator $D_p$ which will be described in detail as below.

We also adopt the $L$-layer CNN structure similar to PatchGANs\cite{pix2pix}\cite{cycle-gan}. However, different from PatchGANs setup where a relatively small stride (1 or 2) is applied, we restrict a special relationship between the stride and other network parameters. Assuming each patch is with  a size of $n \times n$, the convolution kernel at each convolutional layer $l$ has a size of $ k_l \times k_l $, and the stride is set to $s_l$. The following rule will be imposed in our discrimination $D_p$:

\vspace{-0.2cm}

\begin{equation}
  \label{eqn::k-s-p}
  \begin{cases}
  	& k_l = s_l \\
       & n = \prod_{l=1} ^L k_l
  \end{cases}
\end{equation}

\begin{figure}[htb]
\begin{minipage}[b]{.48\linewidth}
  \centering
  \centerline{\includegraphics[width=3.5cm]{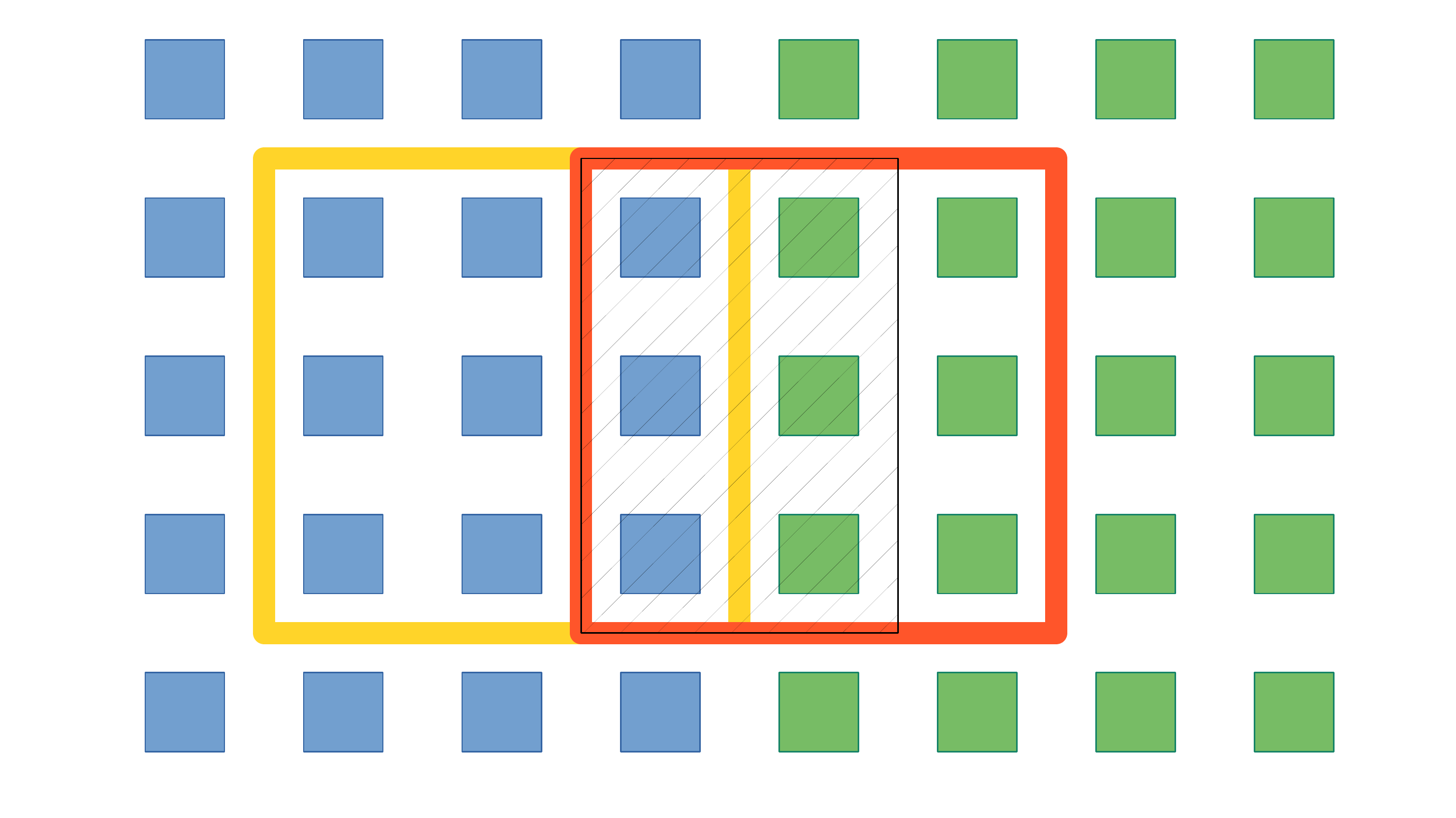}}
  \centerline{(a) stride = 2, kernel\_size = 3}\medskip
\end{minipage}
\hfill
\begin{minipage}[b]{0.48\linewidth}
  \centering
  \centerline{\includegraphics[width=3.5cm]{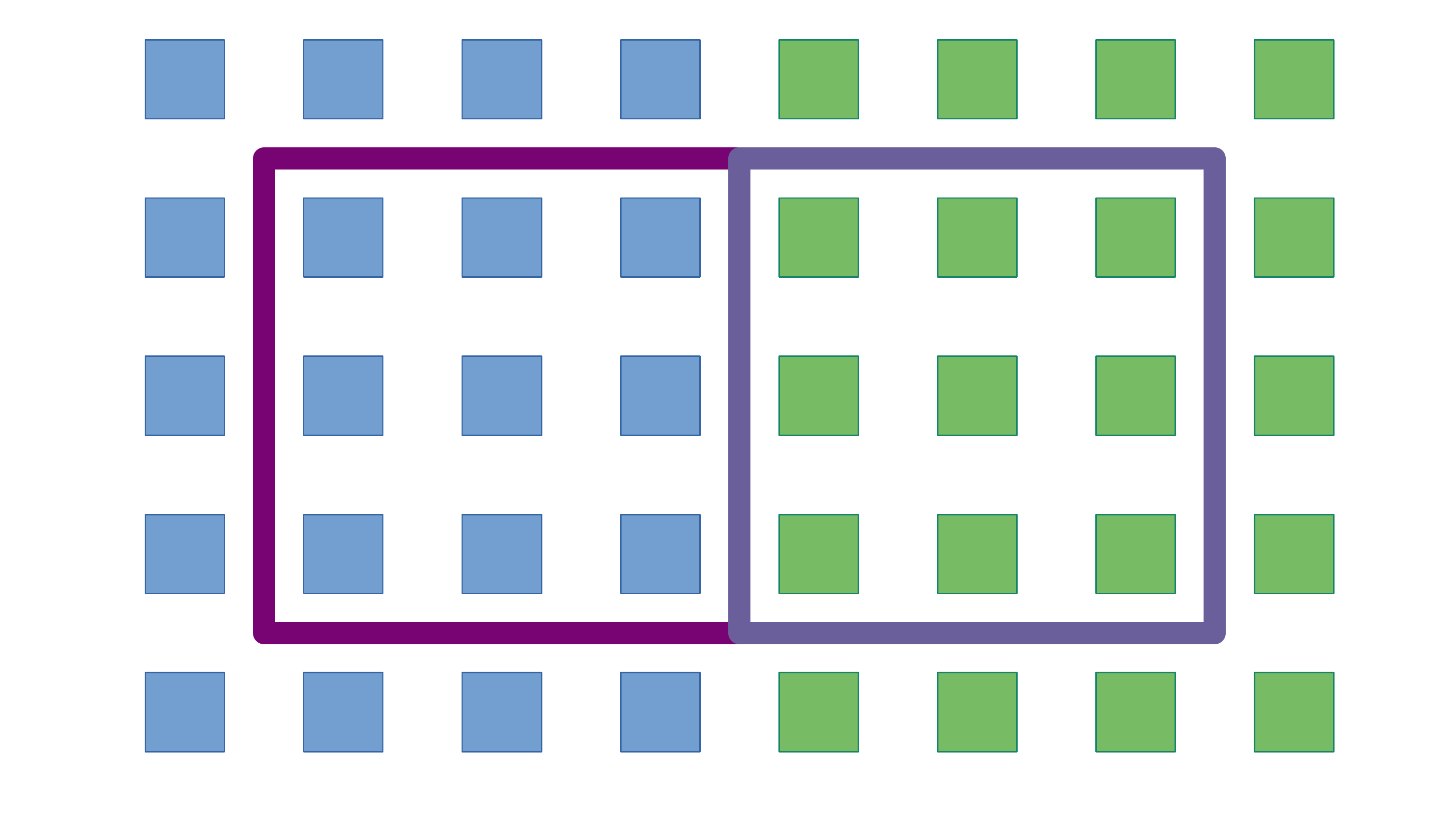}}
  \centerline{(b) stride = kernel\_size = 3}\medskip
\end{minipage}
\vspace{-0.2cm}

  \caption{Principle of the patch discriminator. We set the kernel size and convolutional stride to be a factor of the patch size as in (b) to enable our discriminator on patch permuted images. In contrast, arbitrary setting in standard convolution as in (a) will produce severe artifacts due to discontinuous texture across adjacent patches.}
  \label{fig::discriminator}
\end{figure}

This rule will ensure that any convolution computation in each layer will be performed on one inner patch, and the sliding kernel will never have chance to stretch across different patches. This means our network will only feed forward complete textures from $\psi_i(y_1)$ although discontinuous textures do exist in it. Therefore, unlike PatchGANs which aim to classify overlapping image patches are real or fake, our discriminator classifies the independently patches. Since the network parameters in $Dp$ are relevant to patch size, we name $Dp$ the \emph{patch discriminator} in this paper. The principle of the stride design in $Dp$ is illustrated in Figure.\ref{fig::discriminator}.

\vspace{-0.1cm}

\vspace{-0.5cm}

\begin{table}[htb]
\centering
\begin{tabular}{c|l|c|l}
\toprule
\hline
Operator     & Kernel Shape                       & Stride & Activation   \\
\hline
\emph{conv\_1}      & $3 \times 3 \times 3 \times 256$     & 3   & LeakyReLU        \\
\hline
\emph{conv\_2}      & $3 \times 3 \times 256 \times 512$ & 3   & LeakyReLU        \\
\hline
\emph{conv\_3}      & $1 \times 1 \times 512 \times 1$     & 1   & Sigmoid    \\
\hline
\emph{pooling}      &  Global Avg Pool                              &  -  & None         \\
\hline
\bottomrule
\end{tabular}
\caption{Configuration of proposed patch discriminator}
\label{tab::d-cfg}
\end{table}

\vspace{-0.2cm}

According to the rule above, a $L=3$ discriminator network is implemented in our experiments as depicted in Table.\ref{tab::d-cfg}. Given an input $\psi_i(y_1)$ with size of $nT \times nT$, there will be $T^2$ elements on the output of \emph{conv\_3}, and finally,  a global average pooling is used to vote up the final discriminative result. Besides the ability of compute patch permuted image, $D_p$ can also process natural images such as $G(x_i)$ with no difference with any traditional discriminator since the CNN architecture hasn't been changed. The ability of simultaneously processing multiple patches in ${\psi_i(y_1)}$ and natural images ${G(x_i)}$ by single discriminator makes our network training very efficient.     

\subsection{Objective Function}
\label{sec::method::obj}

Based on the patch discriminator proposed above,  the adversarial loss function used in this paper is defined as follows:

\vspace{-0.3cm}

\begin{equation}
  \label{equ:ori-gan}
\begin{aligned}
 \mathcal{L}_{adv}(G, D_p) =  \mathbb{E}& _{ y \in \mathcal{Y}_{\psi} }  [\log D_p(y)] + \\
 & \mathbb{E}_{x \in \mathcal{X}} [\log(1 - D_p(G(x)))]
\end{aligned}
\end{equation}

An encoder-decoder Generator $G$ similar to \cite{Ronneberger2015U}\cite{pix2pix} is adopted here to learn  the specific stroke style coming from the unique style image $y_1$ by optimizing the adversarial loss. Meanwhile, to preserve the structure information coming from the content image we compute the content loss in a same way as in \cite{Johnson2016Perceptual}. Let  $\phi(x)$ denote 
the feature map of $x$ on the output of $relu1\verb|_|1$ in a $VGG16$ pretrained on ImageNet \cite{11086137620151201}, the content loss is defined as the normalized squared Euclidean distance
between the corresponding feature maps of original images and transferred images:

\vspace{-0.2cm}

\begin{equation}
	\label{eqn::centent-loss}
	\mathcal{L}_{content}(G) = \frac{1}{CHW} \mathbb{E}_{x \in \mathcal{X} }  \parallel \phi(x) - \phi(G(x)) \parallel_2^2
\end{equation}

\noindent
where $C$, $H$ and $W$ denote the channels, height and weight of the feature maps respectively.

The final objective function is defined as follows:

\vspace{-0.2cm}

\begin{equation}
	\label{eqn::final-obj}
	\min_G \max_{Dp} \mathcal{L}(G, D_p)  =  \mathcal{L}_{adv}(G, D_p) + \lambda \mathcal{L}_{content}(G)
\end{equation}
\noindent
where $\lambda$ is a factor to balance the adversarial loss and content loss. By solving the two-player optimization problem in (\ref{eqn::final-obj}), for any given image $x_t$, the transferred image $G(x_t)$ will be able to inherit the stroke style information coming from image $y_1$, as well as preserve its own content information.  

\subsection{Evaluation Criterion}
\label{sec::method::criterion}

Subjective quality evaluation in widely used in recent style transfer methods\cite{li2019learning}\cite{yao2019attention}. In this paper, we propose a quantitative evaluation criterion based on the local texture similarity. Local Binary Pattern(LBP) is used here as our texture descriptor\cite{lbp}. Given two images $a$ and $b$, we firstly compute the LBP features $h_i \in \mathbb{R}^m$, $i = 1, ..., W$, $g_j \in \mathbb{R}^m$, $j = 1, ..., Z$ on randomly cropped patches from $a$ and $b$ respectively, where $m$ denotes the dimension of feature vector on each patch, $W$ and $Z$ denote the number of patches in $a$ and $b$. Our score function is so that defined as follows:

\vspace{-0.2cm}

\begin{equation}
S(a, b) = \frac{1}{Z} \sum_{j=1} ^Z \min_i\parallel h_i - g_j \parallel
\label{eqn::texture-score}
\end{equation}

In (\ref{eqn::texture-score}), the feature of each patch in $a$ will try to find a most similar patch in $b$, in terms of the shortest LBP distance. By average all $Z$ distances for each patch in $a$, the overall texture similarity between $a$ and $b$ can be measured. The LBP descriptor performed a patch level rather than whole image  can effectively prevent the inference of content/structure information coming from style image.

\section{Experiments}

\subsection{Setup}
\label{sec::imple::train}
The P$^2$-GAN network proposed in this paper is trained on PASCAL VOC 2007 dataset\cite{edselc.2-52.0-7795129811520100601}. $N$ = 9.9k images are used as the content images to train our model. Each $\psi_i(y)$ consists of $24 \times 24$  patches with size of $9 \times 9$. We adopt the standard RMSPropOptimizer\cite{bib.rmsprop} to solve the objective function raised in (\ref{eqn::final-obj}). We set $\lambda$ ranging from $1\times10^{-6}$ to $ 1\times10^{-5}$. It takes about 20 minutes to train our model using a specified style on the platform of single NVIDIA GTX 1080 GPU.

\subsection{Quality of Style Transfer}
\label{sec::expr::eval}

\begin{figure}[h]
\vspace{-0.3cm}
    \begin{minipage}[t]{0.138\linewidth}
    \includegraphics[width=1.35cm]{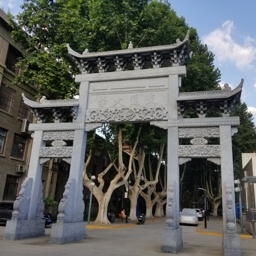}\vspace{1.5pt}
    \includegraphics[width=1.35cm]{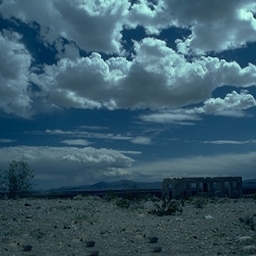}\vspace{1.5pt}
    \includegraphics[width=1.35cm]{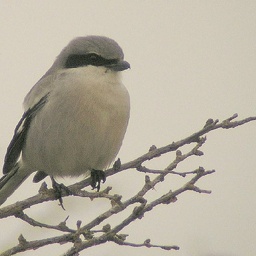}\vspace{1.5pt}
    \includegraphics[width=1.35cm]{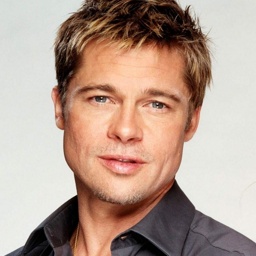}\vspace{1.5pt}
	\centerline{(a) Conte-}
	\centerline{nt}
    \end{minipage}
\hfill
    \begin{minipage}[t]{0.138\linewidth}
    \includegraphics[width=1.35cm]{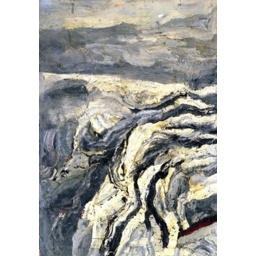}\vspace{1.5pt}
    \includegraphics[width=1.35cm]{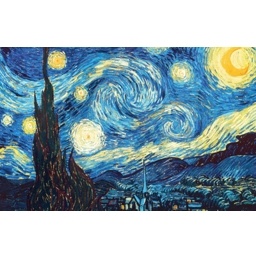}\vspace{1.5pt}
    \includegraphics[width=1.35cm]{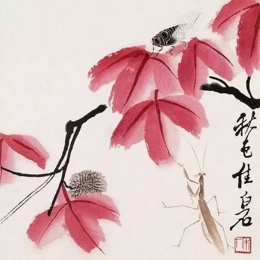}\vspace{1.5pt}
    \includegraphics[width=1.35cm]{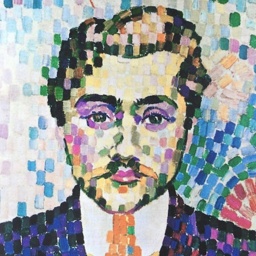}\vspace{1.5pt}
	\centerline{ \ (b) Style}
    \end{minipage}
\hfill
    \begin{minipage}[t]{0.138\linewidth}
    \includegraphics[width=1.35cm]{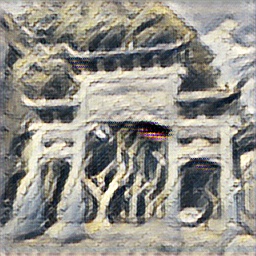}\vspace{1.5pt}
    \includegraphics[width=1.35cm]{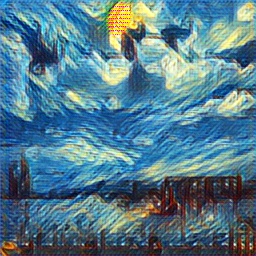}\vspace{1.5pt}
    \includegraphics[width=1.35cm]{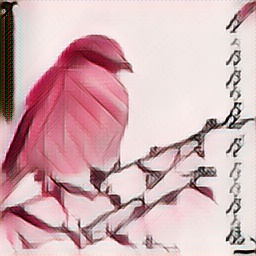}\vspace{1.5pt}
    \includegraphics[width=1.35cm]{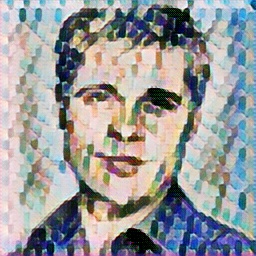}\vspace{1.5pt}
	 \centerline{(c) Johns-}
     \centerline{onNet}
    \end{minipage}
\hfill
    \begin{minipage}[t]{0.138\linewidth}
    \includegraphics[width=1.35cm]{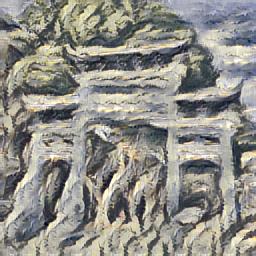}\vspace{1.5pt}
    \includegraphics[width=1.35cm]{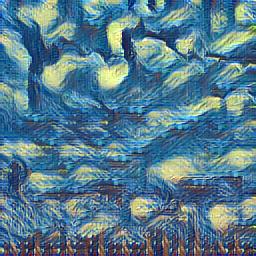}\vspace{1.5pt}
    \includegraphics[width=1.35cm]{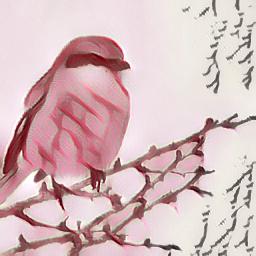}\vspace{1.5pt}
    \includegraphics[width=1.35cm]{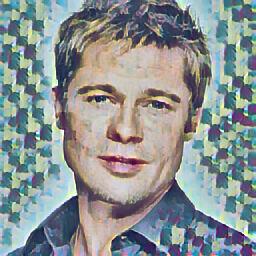}\vspace{1.5pt}
	\centerline{(d) Textu-}
	\centerline{reNetIN}
    \end{minipage}
\hfill
    \begin{minipage}[t]{0.138\linewidth}
    \includegraphics[width=1.35cm]{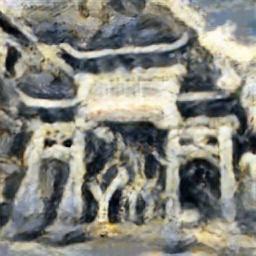}\vspace{1.5pt}
    \includegraphics[width=1.35cm]{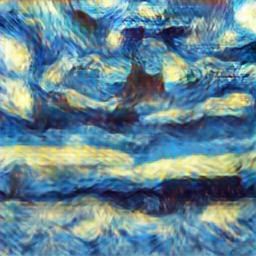}\vspace{1.5pt}
    \includegraphics[width=1.35cm]{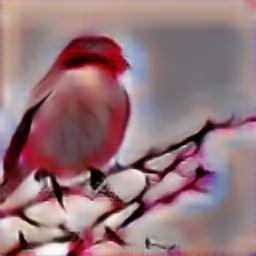}\vspace{1.5pt}
    \includegraphics[width=1.35cm]{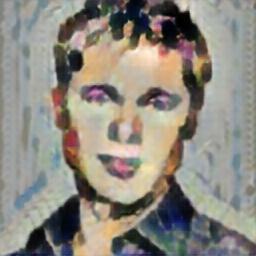}\vspace{1.5pt}
	\leftline{(e) MGAN }
	\centerline{}
    \end{minipage}
\hfill
    \begin{minipage}[t]{0.138\linewidth}
    \includegraphics[width=1.35cm]{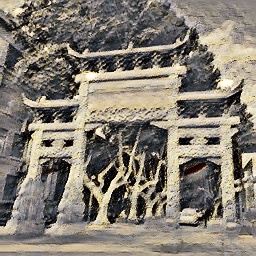}\vspace{1.5pt}
    \includegraphics[width=1.35cm]{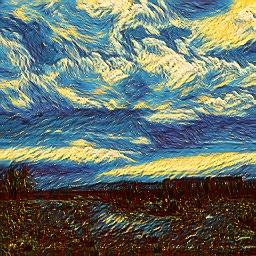}\vspace{1.5pt}
    \includegraphics[width=1.35cm]{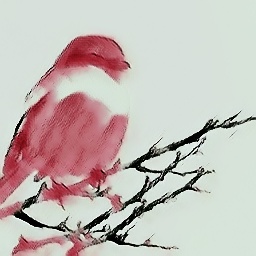}\vspace{1.5pt}
    \includegraphics[width=1.35cm]{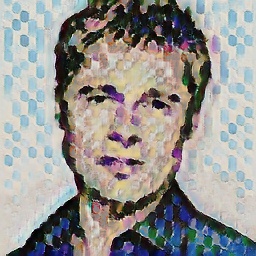}\vspace{1.5pt}
	\leftline{ (f) Ours}\medskip
    \end{minipage}
  \vspace{-0.3cm}
  \caption{Some results of style transfer from single style image. JohnsonNet\cite{Johnson2016Perceptual} and TextureNetIN\cite{stIn} are Gram matrix based methods, MGAN\cite{li2016precomputed} and ours are GAN based methods.}
  \label{fig::comparison}
  \vspace{-0.1cm}
\end{figure}

We compared our method with 3 different style transfer methods including JohnsonNet\cite{Johnson2016Perceptual}, TextureNetIN\cite{stIn} and MGAN\cite{li2016precomputed}. 50 content images are transferred into 5 different styles, including \emph{Mountain No.2} by Jay DeFeo, \emph{The Starry Night} by Van Gogh etc. using different methods respectively. Partial results are shown in Figure.\ref{fig::comparison}.

\begin{figure}[htb]
  \vspace{-0.2cm}
  \centering
  \includegraphics[width=8.0cm]{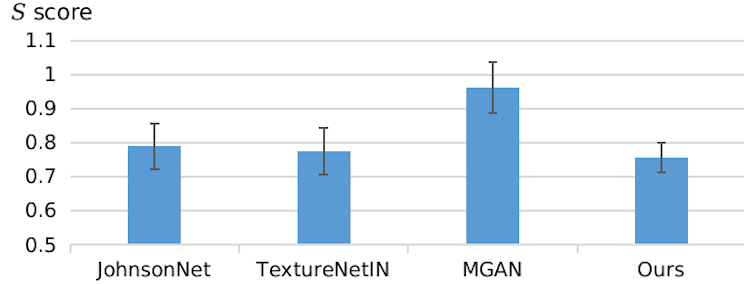}
  \caption{Quantitatively evaluation of the transfer qualities with 4 different methods. The $S$ scores in (\ref{eqn::texture-score}) averaged from 1000 transferred images are plotted.}
  \label{fig::eval::texture}
  \vspace{-0.5cm}
\end{figure}

We also quantitatively compared the transfer quality for all 4 methods using the criterion in (\ref{eqn::texture-score}). The patch size is $32 \times 32$, $W = 1000$, and $Z = 1000$. The mean values and standard deviation of $S$ score from each method are plotted in Figure.\ref{fig::eval::texture}.

The results showed that the MGAN has the highest $S$ score ($0.963 \pm 0.37 $), which means the lowest style similarity between the transferred image and the given one. JohnsonNet and TextureNetIN present the better similarities, that is, $0.790 \pm 0.33$ and $0.776 \pm 0.35$ respectively. In contrast, our method has the lowest $S$ score ($0.756 \pm 0.22$), which indicates the highest texture similarity compared with the given style image. Due to the patch-wise convolution, our method emphasizes more on transferring local texture (stroke style) rather than the global description as done in JohnsonNet and TextureNetIN. Therefore, our method can better avoid the inference of structure information from the training image. 

\subsection{Computational Performance}
\label{sec::expr::t}

To evaluate the computational performance of the proposed method, 1000 images with different resolutions are tested for style transfer, and their online computational time are recorded. The averaged time consumptions per transferred image are summarized in Table. \ref{tab::bench}.

\begin{table}[htb]
\centering
\begin{tabular}{l|c|c|c|c}
\toprule
\hline
\diagbox{Method}{Image size}  & 128              & 256           & 512           & 1024          \\
\hline
JohnsonNet\cite{Johnson2016Perceptual}           & 5.1          & 9.2           & 29.2           & 128           \\
\hline
TextureNetIN\cite{stIn}        & 7.9          & 12.7         & 35.1          & 137          \\
\hline
MGAN\cite{li2016precomputed}                     & 8.2          & 13.2         & 40.7          & - \\
\hline
Ours           & \textbf{3.6} & \textbf{5.6}  & \textbf{13.2}  & \textbf{46.2}  \\
\hline
\bottomrule
\end{tabular}
\vspace{-0.2cm}
\caption{Comparison of averaged computational time (in ms) per style transferring using 4 different methods. The shortest time on each image size is marked in \textbf{bold}.}
\vspace{-0.2cm}

\label{tab::bench}
\end{table}
It can be observed that the time consumptions for all methods are proportional to the image size. Our model overwhelms all other methods and is $2\verb|~|3$ times faster than the others.  This comparison verified the computational efficiency of the proposed network. Specifically, in the processing of $1024 \times 1024$ images, our model achieved a speed of $20+$ FPS which enables real time stylization on HD images.

\section{Conclusion}
In this paper, a novel Patch Permutation GAN (P$^2$-GAN) network that can be efficiently trained by a single style image for the style transfer task is proposed. We use patch permutation to generate multiple training samples, and use a novel patch discriminator to process patch-wise images seamlessly. Our method provides the possibility to precisely simulate the expected stroke style from the customer designated single image, and avoids the difficulty of collecting image set with the same style. Experimental results showed that our method overwhelmed most state-of-the-arts methods in terms of the rendering quality and the computational efficiency.

\bibliographystyle{unsrt}
\bibliography{article}
\end{document}